%% file: main.tex
\documentclass[10pt,twocolumn,letterpaper]{article}

\usepackage[pagenumbers]{cvpr} 

\usepackage{booktabs}
\usepackage{multirow}
\usepackage{graphicx}
\usepackage{array}
\usepackage{gensymb}  
\usepackage{fancyvrb}
\usepackage{lipsum}

\input{preamble}

\makeatletter
\renewcommand{\paragraph}{%
  \@startsection{paragraph}{4}%
  {\z@}{0.25em}{-1em}%
  {\normalfont\normalsize\bfseries}}
\makeatother

\definecolor{cvprblue}{rgb}{0.21,0.49,0.74}
\usepackage[pagebackref,breaklinks,colorlinks,citecolor=cvprblue]{hyperref}

\def\paperID{13186} 
\def\confName{CVPR}
\def\confYear{2024}

\title{Visual Geometry Grounded Deep Structure From Motion}

\author{
Jianyuan Wang$^{1,2}$   
\and 
Nikita Karaev $^{1,2}$    
\and 
Christian Rupprecht$^{1}$    
\and 
David Novotny$^2$    
\and
\\
$^1$Visual Geometry Group, University of Oxford \hspace{5em} 
$^2$Meta AI 
}

\begin{document}

\twocolumn[{%
\renewcommand\twocolumn[1][]{#1}%
\maketitle
\thispagestyle{empty}

\vspace{1em}%
}]

\input{sec/0-abstract}

\begin{figure}[t] 
\includegraphics[width=1.0\linewidth]{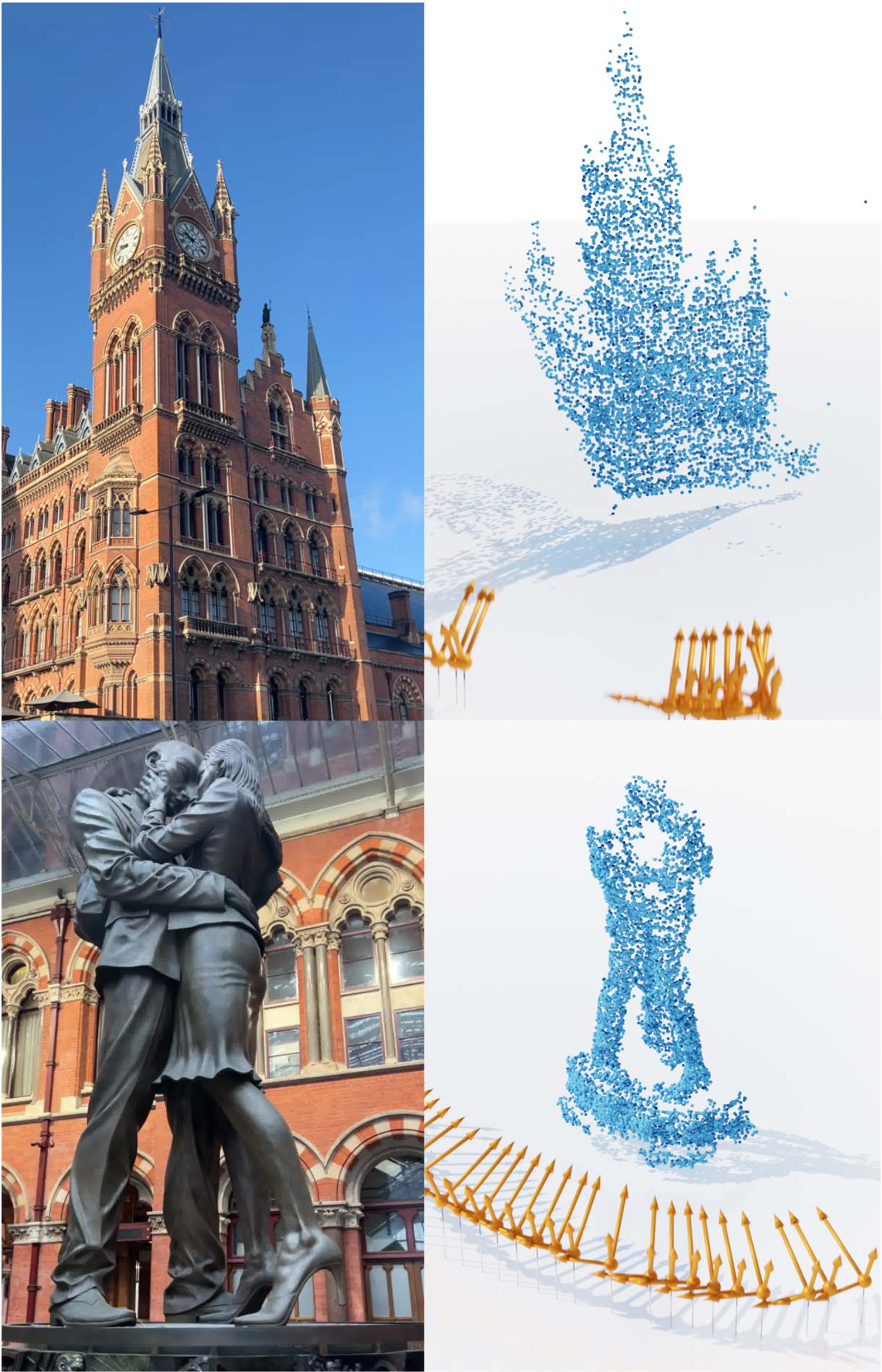}%
\caption{%
\textbf{Reconstruction of In-the-wild Photos with \method},
displaying estimated point clouds (in blue) and cameras (orange).
}\label{fig:splah}
\vspace{-4mm}
\end{figure}

\begin{figure*}[t] 
\centering
\includegraphics[width=1.0\linewidth]{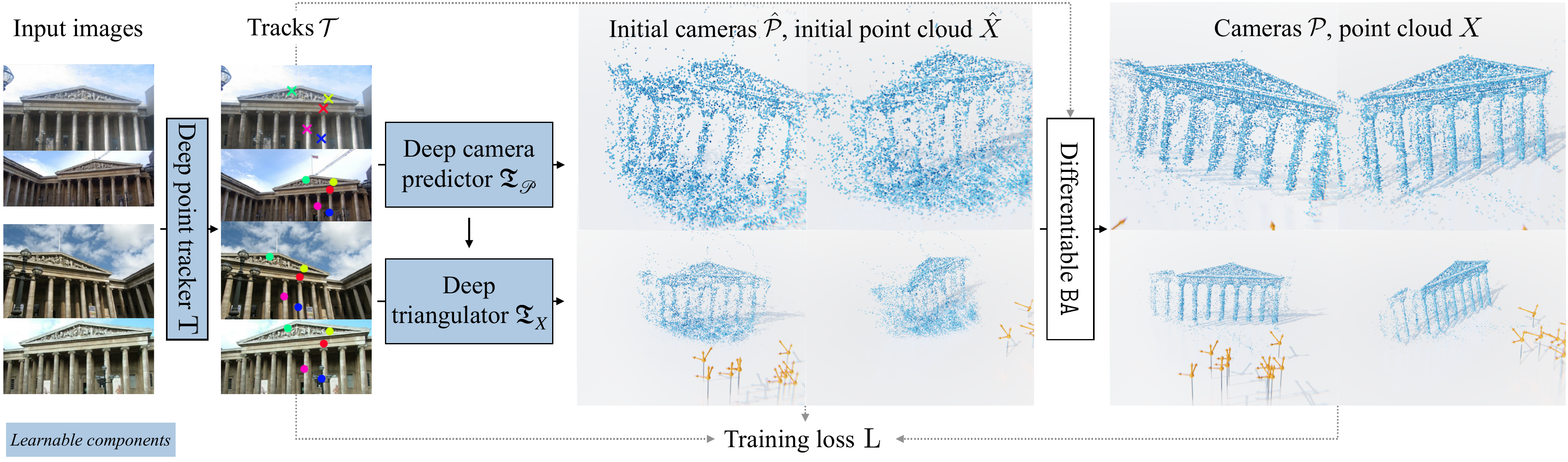}
\caption{%
\textbf{Overview of \method.} Our method extracts 2D tracks from input images, reconstructs cameras using image and track features, initializes a point cloud based on these tracks and camera parameters, and applies a bundle adjustment layer for reconstruction refinement.
The whole framework is fully differentiable and designed for end-to-end training. 
} \label{fig:method_main}
\end{figure*}

\input{sec/1-intro}

\input{sec/2-rw}
\input{sec/3-method}

\input{sec/4-exp}

\vspace{3.5mm}

\input{sec/5-conclusion}


\input{sec/X_suppl}
\newpage

{
    \small
    \bibliographystyle{ieeenat_fullname}
    \bibliography{refs,refs_large}
}


\end{document}

%% file: preamble.tex
\usepackage[dvipsnames]{xcolor}

\newcommand{\method}{VGGSfM\xspace}
\newcommand{\x}{\mathbf{x}}
\newcommand{\y}{\mathbf{y}}
\newcommand{\by}{\mathbf{y}}
\newcommand{\bm}{\mathbf{m}}

\newcommand{\bsigma}{\mathbf{\sigma}}

\newcommand{\setT}{\mathcal{T}}
\newcommand{\setI}{\mathcal{I}}

\newcommand{\setP}{\mathcal{P}}

\newcommand{\tracker}{\texttt{T}}
\newcommand{\badj}{\texttt{BA}}
\newcommand{\tformer}{\mathfrak{T}}
\DeclareMathOperator*{\argmin}{arg\,min}


%% file: sec/0-abstract.tex
\begin{abstract}
Structure-from-motion (SfM) is a long-standing problem in the computer vision community, which aims to reconstruct the camera poses and 3D structure of a scene from a set of unconstrained 2D images. 
Classical frameworks solve this problem in an incremental manner by detecting and matching keypoints, registering images, triangulating 3D points, and conducting bundle adjustment. 
Recent research efforts have predominantly revolved around harnessing the power of deep learning techniques to enhance specific elements (e.g., keypoint matching), but are still based on the original, non-differentiable pipeline.
Instead, we propose a new deep pipeline \method, where each component is fully differentiable and thus can be trained in an end-to-end manner.
To this end, we introduce new mechanisms and simplifications. First, we build on recent advances in deep 2D point tracking to extract reliable pixel-accurate tracks, which eliminates the need for chaining pairwise matches. 
Furthermore, we recover all cameras simultaneously based on the image and track features instead of gradually registering cameras.
Finally, we optimise the cameras and triangulate 3D points via a differentiable bundle adjustment layer.
We attain state-of-the-art performance on three popular datasets, CO3D, IMC Phototourism, and ETH3D.
\end{abstract}

%% file: sec/1-intro.tex
\section{Introduction}
\label{sec:intro}

Reconstructing the camera parameters and the 3D structure of a scene from a set of unconstrained 2D images is a long-standing problem in the computer vision community.
Among many other applications~\cite{ozyecsil2017survey, iglhaut2019structure, westoby2012structure, carrivick2016structure, jiang2020efficient}, it has recently emerged as an important component of learning neural fields \cite{mildenhall2021nerf, kerbl3Dgaussians,chen2022tensorf, jiang2022forge, lin2021barf, wang2021nerfmm}.
The problem is usually solved via the Structure-from-Motion (SfM) framework which estimates the 3D point cloud (Structure) and the parameters of each camera (Motion) in the scene. 
State-of-the-art methods \cite{lindenberger_pixel-perfect_2021,he2023dfsfm} follow the incremental SfM paradigm whose origins can be traced back to the early 2000s \cite{schaffalitzky_multi-view_2002, hartley_multiple_2000}.
It usually begins with a small set of correspondence-rich images as initialization, and gradually adds more views into the reconstruction, through keypoint detection, matching, verification, image registration, triangulation, bundle adjustment (BA), and so on~\cite{schoenberger2016sfm, agarwal2011building,furukawa_towards_2010, heinly_2015_reconstructing}.

Recent research efforts have predominantly revolved around leveraging the power of deep learning techniques to enhance specific elements within the original pipeline while preserving the incremental SfM framework as a whole.
For instance, SuperPoint and SuperGlue \cite{sarlin2020superglue,detone2018self} focus on improving keypoint detection and matching. 
Pixel-perfect SfM \cite{lindenberger_pixel-perfect_2021} proposes deep feature-metric refinement to adjust both keypoints and bundles. 
Detector-free feature matching methods~\cite{sun2021loftr, wang2022matchformer, chen2022aspanformer} bypass early keypoint detection by means of attention, which is powerful in poorly textured scenes.
%
Detector-free SfM \cite{he2023dfsfm} builds a coarse SfM model through quantized detector-free matches and then iteratively refines it with multi-view consistency constraints.
%
These advancements successfully combine deep learning approaches (such as deep feature matching) with well-established hand-engineered components, such as the incremental camera registration of COLMAP~\cite{schoenberger2016sfm}.

The widespread success of end-to-end training warrants the question of what benefits it can bring to long-standing frameworks such as SfM.
Naturally, it is often difficult to assess the merits of new approaches when compared with decades of continuous improvements. 
Nonetheless, in this paper, we answer this question by introducing a fully-differentiable SfM pipeline, dubbed Visual Geometry Grounded Deep Structure From Motion (\method), which trains in an end-to-end manner. 
%
We find that this allows the pipeline to be \textit{simpler} than prior frameworks while achieving better or comparable performance. 
Training end-to-end allows each component to generate outputs that facilitate the task of its successor. 

To build a fully-differentiable pipeline, we make several substantial changes to the SfM procedures and overall obtain better performance. 
Specifically, our model builds on recent advances in deep 2D point tracking \cite{harley2022particle, doersch2022tapvid, doersch2023tapir, karaev2023cotracker} to directly extract reliable pixel-accurate tracks.
This simplifies the correspondence estimation step in traditional SfM, which first estimates pairwise matches and then connects them into tracks.
%
Then, based on the image and track features, \method estimates all cameras jointly via a Transformer~\cite{vaswani_attention_2017}, and subsequently all 3D points. 
Different from \textit{Incremental} SfM, this approach is simpler and easier to differentiate as it does not depend on a discrete, combinatorial correspondence chaining step.
Finally, for bundle adjustment, we replace the commonly employed non-differentiable Ceres solver \cite{Agarwal_Ceres_Solver_2022} with the fully differentiable second-order Theseus solver~\cite{pineda2022theseus}.

Hence, we fuse all the SfM components into a single fully differentiable reconstruction function $f$.
Besides that, in our experiments, we also show that the individual modules perform well in isolation. 
Ultimately, end-to-end training yields another performance improvement, that surpasses the performance of isolated components. 

We evaluate \method for the task of camera pose estimation on the CO3Dv2~\cite{reizenstein_common_2021} and IMC Phototourism~\cite{jin2021image} datasets, and for 3D triangulation on the ETH3D~\cite{schops2017multi} dataset. 
Our method attains strong performance on all benchmarks.
At the same time, we conduct in-the-wild reconstruction to validate the generalization ability of our proposed framework, as shown in \cref{fig:splah}.

%% file: sec/2-rw.tex
\section{Related Work}


\paragraph{Structure from Motion} is a fundamental problem in computer vision and has been investigated for decades~\cite{hartley_multiple_2000, ozyecsil2017survey, oliensis2000critique}.
The classical pipelines usually solve the SfM problem in a global~\cite{moulon2013global, wilson2014robust, cui2015global} or incremental~\cite{snavely2006photo,agarwal2011building, frahm2010building,wu2013towards, schoenberger2016sfm} manner. Both of which are usually based on pairwise image keypoint matching.
Incremental SfM is arguably the most widely adopted strategy (\eg, the popular framework COLMAP~\cite{schoenberger2016sfm}).
Therefore, in the following sections, we refer to incremental SfM as ``classical'' or ``traditional'' SfM.
We defer the discussion of global SfM to the supplementary. 

Traditional SfM frameworks often start by detecting keypoints and feature descriptors~\cite{lowe_object_1999,lowe_distinctive_2004-1, bay_speeded-up_2008, matas_robust_2002}.
They then search for image pairs with overlapping frusta by matching these keypoints across different images (\eg, with a nearest-neighbour search)~\cite{lou2012matchminer, agarwal2011building, schoenberger2016sfm}.
These image pairs are further verified via two-view epipolar geometry or homography~\cite{hartley_multiple_2000} through RANSAC~\cite{fischler1981random}.
Then, a pair or a small set of images is carefully selected for initialization.
New images are gradually registered by solving the Perspective-$n$-Point (PnP) problem~\cite{lu2018review}, followed by triangulating 3D points, and bundle adjustment~\cite{triggs_bundle_2000}. 
This process is iterated until all the frames are either registered or discarded. 
2D correspondences (multi-view tracks) are the basis of the whole process, however, they are usually simply constructed by chaining two-view matches~\cite{schoenberger2016sfm}.


Many deep-learning approaches have been proposed to enhance this framework. For example, \cite{yi_lift_2016, detone2018self, tyszkiewicz2020disk} provide better keypoint detection and ~\cite{sarlin2020superglue, chen2021learning, lindenberger2023lightglue, shi2022clustergnn, jiang2021cotr} focus on matching. Furthermore, detector-free matching methods~\cite{sun2021loftr, wang2022matchformer, chen2022aspanformer} propose to avoid sparse keypoint detection by building semi-dense matches via self and cross attention. Some studies improve the performance of RANSAC by making it trainable~\cite{brachmann2017dsac,brachmann2019neural, wei2023generalized}. Recent state-of-the-art methods are PixSfM~\cite{lindenberger_pixel-perfect_2021} and the concurrent Detector-free SfM (DFSfM)~\cite{he2023dfsfm}. PixSfM refines the tracks and structure estimated by COLMAP through feature-metric keypoint adjustment and feature-metric bundle adjustment. Detector-free SfM proposes to first build a coarse SfM model using detector-free matches and COLMAP (or other frameworks), and then to iteratively refine the tracks and the structure of the coarse model by enforcing multi-view consistency. 

Recently, fully differentiable SfM pipelines have also been explored. They usually use deep neural networks to regress camera poses and depths~\cite{zhou2017unsupervised,ummenhofer2017demon,tang2018ba,wei2020deepsfm,wang2021deep, teed2018deepv2d, teed2021droid}. Although using an approximation of bundle adjustment~\cite{tang2018ba, wei2020deepsfm,teed2021droid}, these methods suffer from limited generalizability and scalability (very few input frames)~\cite{zhou2017unsupervised, ummenhofer2017demon, tang2018ba, wang2021deep, wei2020deepsfm}, or rely on temporal relationship~\cite{teed2018deepv2d, teed2021droid}. Meanwhile, some methods are category-specific~\cite{wu2020unsupervised,ma2022virtual,wu2023magicpony}. The recent efforts on deep camera pose estimation can scale up to more than 50 frames, but they do not reconstruct the scene~\cite{zhang2022relpose,sinha2023sparsepose, wang2023posediffusion, lin2023relposepp}. 

\paragraph{Point tracking.} Since \method proposes a novel point tracker, next, we review recent advances in this field. 
Inspired by the optical-flow architecture of RAFT~\cite{teed2020raft}, PIPs~\cite{harley2022particle} revisited point tracking, a task related to Particle Video~\cite{sand2008particle}, and proposed a highly accurate tracker of isolated points in a video.
TAP-Vid~\cite{doersch2022tapvid} (\ie, \textit{``Tracking Any Point''}) introduced a benchmark for point tracking and a baseline model, which was later improved in TAPIR~\cite{doersch2023tapir} by integrating the iterative update mechanism from PIPs.
PointOdyssey~\cite{zheng2023pointodyssey} simplified PIPs and proposed a benchmark for the long-term version of point tracking. CoTracker~\cite{karaev2023cotracker} closed the gap between single point tracking and dense Optical Flow with joint point tracking.
However, these works are designed for videos, \ie temporally-ordered sequences of frames.
In our point tracker, given the input frames are unordered, we do not assume a temporal relationship between input frames.
We therefore process all frames jointly, avoiding windowed inference of \cite{karaev2023cotracker}.
Since SfM relies on highly accurate correspondences, our tracks are further refined in a coarse-to-fine manner to achieve sub-pixel accuracy. 

%% file: sec/3-method.tex
\section{Method}


In this section, we describe the components of \method and how they are composed in a fully differentiable pipeline.
An overview of our framework is shown in \cref{fig:method_main}.

\paragraph{Problem setting}
Given a set of free-form images observing a scene, 
\method estimates their corresponding camera parameters and the 3D scene shape represented as a point cloud. Formally, given a tuple $\setI = \big(I_1, ..., I_{N_I}\big)$ of $N_I \in \mathbb{N}$ RGB images $I_i \in \mathbb{R}^{3 \times H \times W}$, \method estimates the corresponding camera projection matrices $\setP = \big( P_1, ..., P_{N_I} \big | P_i \subset \mathbb{R}^{3 \times 4} \big)$ and the scene cloud $X = \{\x^j\}_{j=1}^{N_\x}$ of $N_\x \in \mathbb{N}$ 3D points $\x^j \in \mathbb{R}^3$.
Each projection matrix $P_i$ consists of extrinsics (pose) $g_i \in \mathbb{SE}(3)$ and intrinsics $K_i \in \mathbb{R}^{3 \times 3}$.

A 3D point $\x^j$ can be projected to the $i$-th camera yielding a 2D screen coordinate $\y_i^j = P_i(\x^j) \sim \lambda K_i \hat{\x}_i^j; \lambda \in \mathbb{R}_+$, where $\hat{\x}_i^j = g_i \x^j$ is the world coordinate $\x^j$ expressed in view-coordinates of the $i$-th camera. 
The projection of the point $\x^j$ to all input cameras is a \textit{track} $T^j = \big( (y_1^j, v_1^j) ..., (y^j_{N_I}, v^j_{N_I}) \big )$ consisting of $N_I$ matching 2D points $\y_i^j \in \mathbb{R}^2$, and their corresponding binary indicators $v_i^j \in \{0, 1\}$ denoting visibility of the $j$-th point in the $i$-th camera.
We denote $\setT_i = \{T^1_i, ..., T_i^{N_T} \}$ as the set of all tracks $T^j_i$ in the $i$-th camera. 

\subsection{Method overview}
\method implements SfM via a single function $f_\theta$
\vspace{-1mm}
\begin{equation}
f_\theta(\setI) = \setP, X
\vspace{-1mm}
\end{equation}
accepting the set of $N_I$ scene images $\setI$ and outputting the camera parameters $\setP$ and the scene point cloud $X$.
Importantly, $f_\theta$ is fully differentiable and, as such, its parameters $\theta$ are learned by minimizing the training loss $\mathcal{L}$:
\vspace{-1mm}
\begin{equation} \label{eq:main_loss}
\theta^\star
=
\argmin_\theta
\sum_{s=1}^{S} \mathcal{L}(f_\theta(\setI_s), \setP^\star_s, \setT^\star_s, X_s^\star),
\vspace{-1mm}
\end{equation}
summing over $S \in \mathbb{N}$ training image sets $\setI_s$ annotated with ground-truth cameras $\setP_s^\star$, tracks $\setT_s^\star$, and point clouds $X_s^\star$.
We defer the details of $\mathcal{L}$ to \cref{sec:method_details} and, in the following paragraphs, discuss the architecture of $f_\theta$.


\paragraph{The reconstruction function}
Following traditional SfM \cite{schoenberger2016sfm},
\method decomposes the reconstruction function $f_\theta$ into four seamless stages: 1) point tracking $\tracker$, 2) initial camera estimator $\tformer_{\setP}$, 3) triangulator $\tformer_{X}$ and, 4) Bundle Adjustment $\badj$, as follows:
\begin{equation}
\vspace{-1mm}
\begin{aligned}
\setT &= \tracker(\setI) \\
\hat{\setP} &= \tformer_{\setP}(\setI, \setT) \\
\hat{X} &= \tformer_X(\setT, \hat{\setP}) \\
\setP, X &= \badj(\setT, \hat{\setP}, \hat{X}). \\
\end{aligned}
\end{equation}
The tracker $\tracker$ estimates 2D tracks $\setT$ given input images $\setI$. Subsequently, $\tformer_{\setP}$ and $\tformer_{X}$ provide initial cameras $\hat{\setP}$ and an initial point cloud $\hat{X}$ respectively.
Finally, $\badj$ enhances accuracy by refining the cameras and 3D points together.

\subsection{Tracking} \label{sec:tracking}
Establishing precise 2D correspondences is important for achieving accurate 3D reconstruction. Traditionally, SfM frameworks first estimate pairwise image-to-image correspondences that are later chained into multi-image tracks $T$~\cite{lindenberger_pixel-perfect_2021, schoenberger2016sfm,he2023dfsfm}.
Here, typically only point-pair matching benefits from learned components~\cite{detone2018self, tyszkiewicz2020disk,sarlin2020superglue,lindenberger2023lightglue}, while the chaining of pairwise correspondences remains a hand-engineered process.

Instead, \method \emph{significantly simplifies SfM correspondence tracking} by employing a deep feed-forward tracking function. It accepts a collection of images and directly outputs a set of reliable point trajectories across all images. 
We achieve this by exploiting recent advances in video point tracking methods~\cite{harley2022particle, doersch2022tapvid, doersch2023tapir, karaev2023cotracker}. 
%
Although developed for video-point tracking, these methods are inherently appropriate for SfM which requires a very compact set of highly accurate tracks (\eg, dense optical flow is too memory-demanding).
Furthermore, point trackers mitigate the potential errors (sometimes called drift) caused by chaining of pairwise matches. 
However, as we describe below, our design differs from video trackers because SfM, which accepts free-form images, cannot assume temporal smoothness or ordering, and requires sub-pixel accuracy.

\begin{figure}[htbp] 
\includegraphics[width=1.0\linewidth]{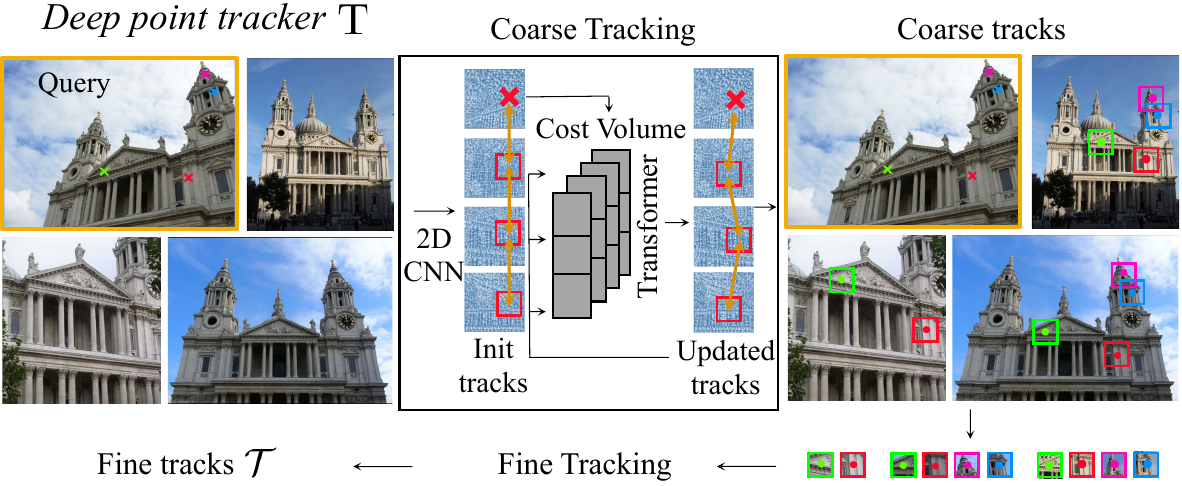}
\caption{\textbf{Architecture of Tracker $\tracker$.}
We adopt a coarse-to-fine design for the tracker. The coarse tracker first locates the approximate positions of corresponding points, and the fine tracker then refines these initial predictions.
}
\label{fig:archtrack}
\end{figure}

\paragraph{Tracker architecture}
The design of our tracker $\tracker$ follows \cite{harley2022particle,karaev2023cotracker}, and is illustrated in \cref{fig:archtrack}.
More specifically, given $N_T$ query points $\{\hat{\y}^1_i, ... \hat{\y}^{N_T}_i\}$ in a frame $I_i$, we bilinearly sample their corresponding query descriptors $\{\bm^1_i,..., \bm^{N_T}_i\}$ from image feature maps output by a 2D CNN.
Then, each query descriptor is correlated with the feature maps of all $N_I$ frames at different spatial resolutions, which constructs a cost-volume pyramid. 
Flattening the latter yields tokens $V \in \mathbb{R}^{N_T \times N_I \times C}$, where $C$ is the total number of elements in the cost-volume pyramid.
%
%
Feeding the tokens to a Transformer, we obtain tracks $\setT = \{T^j\}_{j=1}^{N_T}$.
Recall that each track $T^j$ comprises the set of $N_I$ tracked 2D locations $\y_i^j$ together with predicted visibility indicators $v_i^j$. 

It is worth noting that, differently from \cite{harley2022particle,karaev2023cotracker}, our tracker does not assume temporal continuity.
Therefore, we avoid the sliding window approach and, instead, attend to all the input frames together.
Furthermore, unlike in \cite{karaev2023cotracker}, we predict each track independently of others.
This allows to track a larger number of points at test time leading to increased density of reconstructed point clouds. 




\paragraph{Predicting tracking confidence}
In SfM, it is crucial to filter out any outlier correspondences as they can negatively impact the subsequent reconstruction stages.
To this end, we enhance the tracker with the ability to estimate confidence for each track-point prediction.

%
More specifically, we leverage the aleatoric uncertainty \cite{kendall_what_2017,novotny_learning_2017} model which predicts variance $\bsigma_i^j$ together with each 2D track point $\y_i^j$, so that the resulting normal distribution $\mathcal{N}(\y_i^{j\star} | \by_i^j, \bsigma_i^j)$ tightly peaks around each ground-truth 2D track point $\y_i^{j\star}$.
Hence, during training, the $\ell_1$/$\ell_2$ loss, originally used in video point tracking, is replaced with the (negated) logarithm of the latter probability evaluated at each ground truth point $\y_i^{j\star}$.
Once trained, the confidence measure $1 / \sigma_i^j$ is proportional to the inverse of the predicted variance.
In practice, we assume a diagonal covariance matrix resulting in horizontal and vertical uncertainties $\bsigma_i^j$.

\paragraph{Coarse-to-fine tracking}
Moreover, since SfM requires highly accurate (pixel or sub-pixel level) correspondences, we employ a coarse-to-fine point-tracking strategy.
As described above, we first coarsely track image points using feature maps that fully cover the input images $I$.
Then, we form $P \times P$ patches by cropping input images around the coarse point estimates and execute the tracking again to obtain a sub-pixel estimate.
Recall that, differently from the chained matching of traditional SfM, our tracker is fully differentiable.
This enables back-propagating the gradient of the training loss $\mathcal{L}$ through the whole framework to the tracker parameters.
This reinforces the synergy between the tracking and the ensuing stages, which are described next.

\subsection{Learnable camera \& point initialization} \label{sec:initializers}
As discussed above, a traditional SfM pipeline \cite{schoenberger2016sfm,lindenberger_pixel-perfect_2021} usually relies on an an incremental loop, which often initializes with a correspondence-rich image pair, gradually registers new frames, enlarges the point cloud, and conducts joint optimization (\eg, BA). 
%
However, although the framework has been fortified in robustness and accuracy through decades of improvements, this cumulative process has inevitably led to increased complexity. 
Furthermore, Incremental SfM is largely non-differentiable which complicates end-to-end learning from annotated data. 

Thus, in pursuit of simplicity and differentiability, our method departs from the classical SfM scheme. 
Inspired by recent advances in deep camera pose estimators \cite{wang2023posediffusion,lin2023relposepp,zhang2022relpose}, we propose to initialize the cameras and the point cloud with a pair of deep Transformer \cite{vaswani_attention_2017} networks.
Importantly, we register all cameras and reconstruct all scene points collectively in a non-incremental differentiable fashion.
%

\paragraph{Learnable camera initializer}
The predictor of initial cameras $\hat{\setP}$ is implemented as a deep Transformer architecture $\tformer_{\setP}$:
\begin{equation} \label{eq:pose_transformer}
\hat{\setP}
=
\tformer_{\setP}(
\{ \phi(I_i) | I_i \in \setI \},
\{
    d^{\setP}(y_i^j) | \forall T_i \in \setT ~ \forall y_i^j \in T_i
\}
).
\end{equation}
It accepts a set of tokens comprising global ResNet50 \cite{he_deep_2015} features $\phi(I_i)$ of input images $\setI$,
and the set of descriptors $d^{\setP}(\y_i^j)$ of track points $\y_i^j \in T_i \in \setT$. 
Here, each track descriptor is output by an auxiliary branch of the tracker $\tracker$.
Given these inputs, $\tformer_{\setP}$ first applies cross-attention between the global image feature (query) and the track-descriptor (key-value) pairs yielding $N_I = |\setI|$ tokens per scene.
The output of cross-attention is then concatenated with an embedding of a preliminary camera estimated by the 8-point algorithm taking the correspondences between track points $\y_i^j$ as input.
Finally, we feed this concatenation to a Transformer trunk resulting in the initial cameras $\hat{\setP}$.

\paragraph{Learnable triangulation}
Given initial cameras $\hat{\setP}$ and 2D tracks $\setT$, the triangulator outputs the initial point cloud $\hat{X}$.
Similar to the camera predictor, the triangulator is a Transformer $\tformer_X$
\begin{equation}
\hat{X}
=
\tformer_X(
\{
    d^X(y_i^j) | \forall T_i \in \setT \forall y_i^j \in T_i
\}
)
\end{equation}
accepting descriptors $d^X(\y_i^j)$ each comprising a tracker feature, and a positional harmonic embedding \cite{mildenhall_nerf_2020} of points $\bar{\x}^j \in \bar{X}$ from a preliminary point cloud $\bar{X}$.
The preliminary point cloud is formed via closed-form multi-view Direct Linear Transform (DLT) 3D triangulation \cite{hartley_multiple_2000} of the tracks $\setT$ given the initial cameras $\hat{\setP}$.
Please refer to the supplementary for a detailed description of both initializers.

\subsection{Bundle adjustment} \label{sec:ba}
Given the tracks $\setT$ (\cref{sec:tracking}), initial cameras $\hat{\setP}$, and the initial point cloud $\hat{X}$ (\cref{sec:initializers}), 
Bundle Adjustment $\badj$ minimizes the reprojection loss $\mathcal{L}_\text{BA}$\cite{hartley_multiple_2000,schoenberger2016sfm,agarwal2011building, agarwal2010bundle}:
\begin{equation} \label{eq:ba_repro_loss}
\begin{aligned}
X, \setP
&=
\badj(\setT, \hat{X}, \hat{\setP})
=
\argmin_{X, \setP} \mathcal{L}_\text{BA} \\
\mathcal{L}_\text{BA} 
&=
\sum_{i=1}^{N_I} \sum_{j=1}^{N_\x} v_i^j \| P_i(\x^j) - y_i^j \|,
\end{aligned}
\end{equation}
summing over all reprojection errors $\| P_i(\x^j) - \y_i^j \|$ each comprising the distance between the projection $P_i(\x^j)$ of the point cloud  $\x^j \in X$ to camera $P_i \in \setP$, and the $i$-th 2D point $\y_i^j \in T^j$ of the track $T^j$.
Additionally, the error terms are filtered out if the corresponding points have low visibility, low confidence, or do not fit the geometric constraints defined by~\cite{schoenberger2016sfm}. 
Points with large reprojections errors are also filtered~\cite{snavely2006photo, wu2013towards, schoenberger2016sfm}. More details are provided in the supplementary material. 

\paragraph{Differentiable Levenberg-Marquardt}
Following common practice \cite{schoenberger2016sfm,lindenberger_pixel-perfect_2021}, we minimize \cref{eq:ba_repro_loss} with second-order Levenberg-Marquardt (LM) optimizer~\cite{more2006levenberg}.
However, optimizing the main training loss (\cref{eq:main_loss}) via backpropagation requires differentiability of \cref{eq:ba_repro_loss} which is non-trivial. 
Therefore, we leverage the recently proposed Theseus library~\cite{pineda2022theseus} which exploits the implicit function theorem to backpropagate through deep networks with nested optimization loops.

\begin{figure*}[htbp] 
\includegraphics[width=1.0\linewidth]{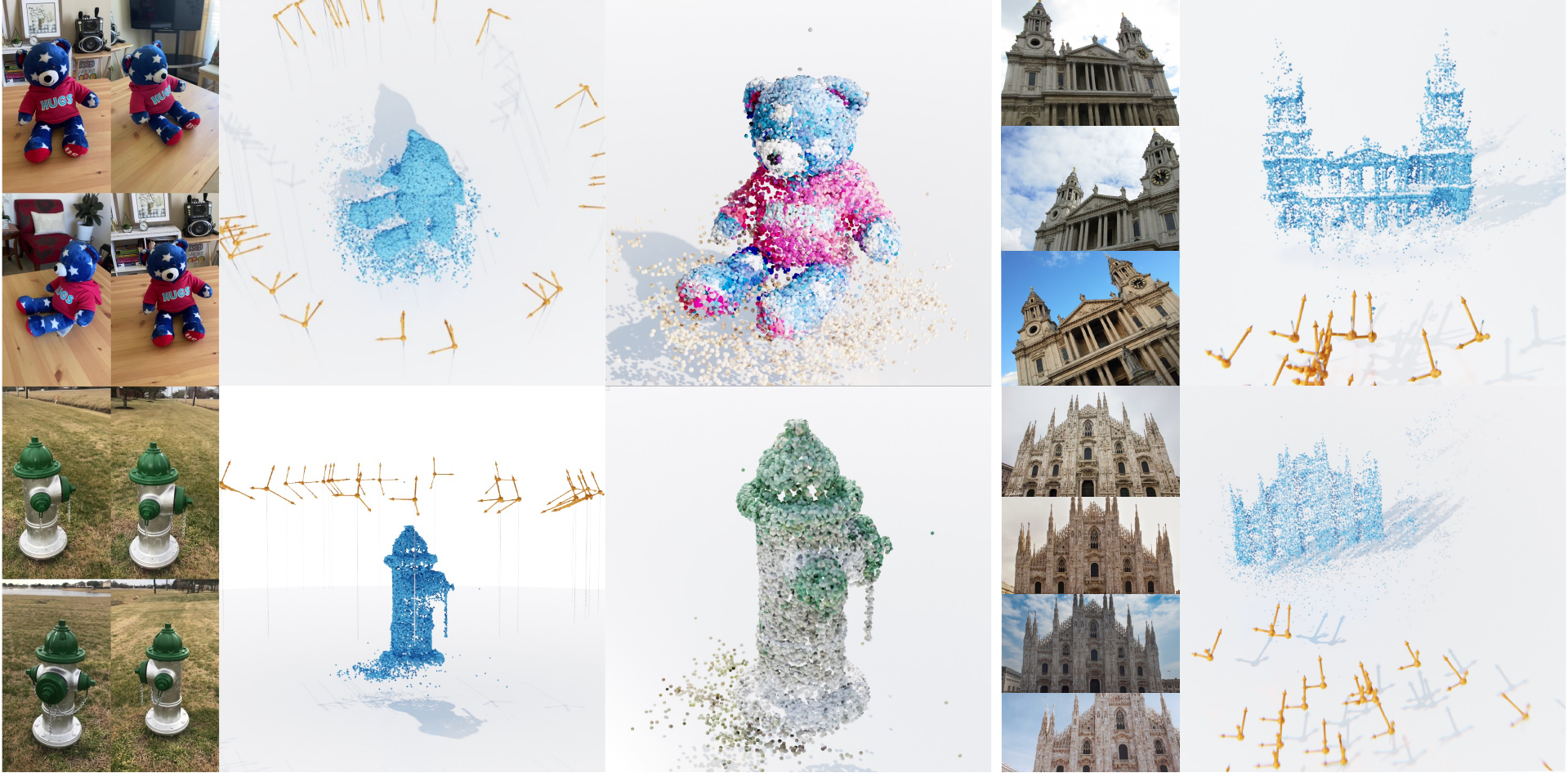}%
\caption{\textbf{Camera and point-cloud reconstructions} of \method on Co3D (left) and IMC Phototourism (right). 
}\label{fig:denserec}
\end{figure*}

\subsection{Method details} \label{sec:method_details}

\paragraph{Camera parameterization}
Each camera pose $P \in \setP$ is parameterized with 8-degrees of freedom: the quaternion $q(R) \in \mathbb{R}^4$ of the rotation $R \in \mathbb{SO}(3)$ and the translation $\mathbf{t} \in \mathbb{R}^3$ components of $P$'s extrinsics $g \in \mathbb{SE}(3)$, and the logarithm $\ln(\mathfrak{f}) \in \mathbb{R}$ of the camera focal length $\mathfrak{f} \in \mathbb{R}^+$.
Given these values, the $3 \times 4$ pose matrix is defined as $P = K R [\mathbb{I}_3 | \mathbf{t}]$, with the calibration matrix
$K = [\mathfrak{f}, 0, p_x; 0, \mathfrak{f}, p_y; 0, 0, 1] \subset \mathbb{R}^{3 \times 3}$ (row major order).
Following standard practice \cite{schoenberger2016sfm,lindenberger_pixel-perfect_2021}, we set the principal-point coordinates $p_x, p_y \in \mathbb{R}$ to the center of the image. 

\paragraph{Training loss}
The training loss $\mathcal{L}$ (\cref{eq:main_loss}) is defined as:
{
\thinmuskip=1mu
\medmuskip=2mu
\thickmuskip=3mu
\begin{equation} \label{eq:detailed_loss}
\begin{aligned}
&\mathcal{L}(f_\theta(\setI),\setP^\star,\setT^\star,X^\star)
=
\sum_{j=1}^{N_T} | \x^{\star j} - \x^{j} |_\epsilon + | \x^{\star j} - \hat{\x}^{j} |_\epsilon +
\\
&
+ \sum_{i=1}^{N_I} e_\setP( P^\star_i, P_i) + e_\setP( P^\star_i, \hat{P}_i) 
- \sum_{i=1}^{N_I}\sum_{j=1}^{N_T} \log \mathcal{N}(\y^{j\star}_i | \y_i^j, \sigma_i^j)
\end{aligned}
\end{equation}
}
Here,
$| \x^{\star j} - \x^{j,t} |_\epsilon$ and $| \x^{\star j} - \hat{\x}^{j} |_\epsilon$ evaluate 
the $\epsilon$-thresholded pseudo-Huber loss $| \cdot |_\epsilon$ \cite{charbonnier_deterministic_1997}
between the ground truth 3D points $\x^{\star j}$ and the initial and BA-refined 3D points  $\hat{\x}^j \in X$, $\x^j \in \hat{X}$ respectively.
The camera errors $e_\setP( P^\star_i, P_i)$ and $e_\setP( P^\star_i, \hat{P}_i)$ compare the predicted initial pose $\hat{P}_i \in \hat{\setP}$ and bundle-adjusted camera pose $P_i \in \setP$ to the ground truth camera annotation $P^\star_i \in \setP^\star_i$.
Here, $e_\setP(P, P')$ is defined as the Huber-loss $| \cdot |_\epsilon$ between the parameterizations of poses $P$, $P'$.
Finally, $\log \mathcal{N}(\y^{j\star}_i | \y_i^j, \sigma_i^j)$ computes the likelihood of a ground-truth track point $\y^{j\star}_i \in T^\star_i$ under a probabilistic track-point estimate defined by a 2D gaussian with mean and variance predictions $\y_i^j$ and $\sigma_i^j$ respectively (i.e. the aleatoric uncertainty model described in \cref{sec:tracking}).

%% file: sec/4-exp.tex
\section{Experiments}\label{sec:exp}
In this section, we first introduce the datasets together with the protocols for training and evaluation. 
%
Then, we provide comparison to existing methods and ablation studies. 

\begin{figure*}[htbp] 
\includegraphics[width=1.01\linewidth]{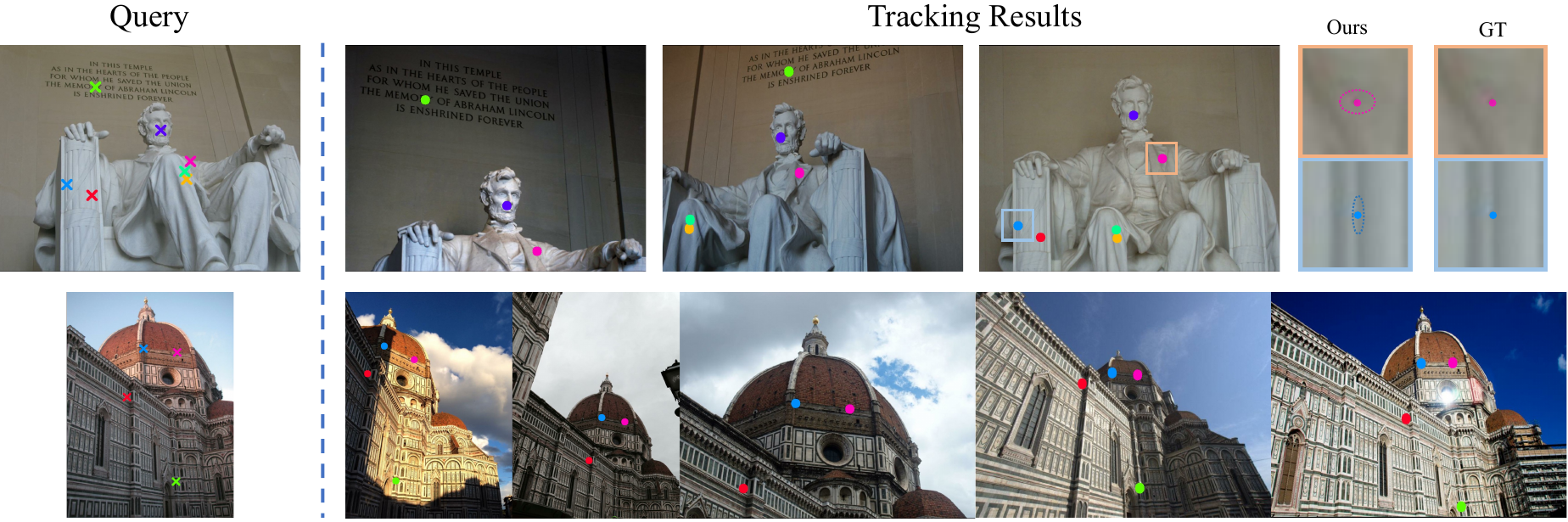}
\caption{
\textbf{Qualitative Evaluation of Tracking.}
In each row, the left-most frame contains the query image with query points (crosses).
The predicted track points $\y_i^j$ (dots) are shown to the right. 
The top-right part also highlights our track-point confidence predictions (described in \cref{sec:tracking}), illustrated as ellipses with extent proportional to the predicted variance $\bsigma_i^j$.
Note how the uncertainty corresponds to an expectation, \eg, a keypoint covering a vertical stripe has higher uncertainty along the $y$-axis.
} \label{fig:track_qual}
\end{figure*}

\begin{table*}[t]
{\centering
    \small
    \resizebox{0.95\linewidth}{!}{%
        \begin{tabular}{c|cccccc|cc}
        \toprule
        {Co3D Dataset}     & \multicolumn{2}{c|}{Incremental SfM} & \multicolumn{6}{c}{Deep}    \\
        \midrule
        Method  & COLMAP (SP + SG)  &  {PixSfM (SP + SG)} & {RelPose\cite{zhang2022relpose}}  & {PoseReg\cite{wang2023posediffusion}}    & RelPose++\cite{lin2023relposepp} & {PoseDiffusion\cite{wang2023posediffusion}}  & Ours w/o Joint & Ours \\
        \midrule
        RRE@15$\degree$  & 31.6 & 33.7 & 57.1 & 53.2    & {82.3}  & {80.5} & \underline{88.2} & \textbf{92.1}   \\
        RTE@15$\degree$  & 27.3 & 32.9 & - & 49.1   & 77.2  & {79.8} & \underline{83.4} & \textbf{88.3} \\
        AUC@30$\degree$ & 25.3 & 30.1  & - &  45.0    & 65.1 & {66.5} &  \underline{70.7} &  \textbf{74.0} \\
        \bottomrule
        \end{tabular}
    }
  \caption{
      \textbf{Camera Pose Estimation on Co3D,} where the proposed method outperforms previous methods by a large margin.  Ours w/o Joint indicates the variant of our framework without training all components jointly.
      \label{tab:Co3Dpose}
  }    
}
\end{table*}

\paragraph{Datasets.} Following prior work~\cite{wang2023posediffusion, lindenberger_pixel-perfect_2021, he2023dfsfm}, we evaluate camera pose estimation on Co3Dv2~\cite{reizenstein_common_2021} and IMC Phototourism datasets~\cite{jin2021image}, and 3D triangulation on ETH3D~\cite{schops2017multi}.
Co3D is an object-centric dataset comprising turnable-style videos from 51 categories of MS COCO~\cite{lin_microsoft_2014}.
IMC Phototourism, provided by Image Matching Challenge~\cite{jin2021image}, contains 8 testing scenes and 3 validation scenes of famous landmarks.
Generally, the Co3D scenes have much wider baselines, making them challenging for traditional frameworks such as COLMAP, while the IMC samples often have sufficiently overlapping fursta, which is where COLMAP excels.
ETH3D provides highly accurate point clouds (captured by laser scanner) for 13 indoor and outdoor scenes, and hence is suitable for the evaluation of triangulation.

%


\paragraph{Training.}
For the model evaluated on IMC Phototourism and ETH3D, we 
follow the protocol of ~\cite{sarlin2020superglue, lindenberger_pixel-perfect_2021, he2023dfsfm} and train on the MegaDepth dataset~\cite{li2018megadepth}.
MegaDepth contains 1M crowd-sourced images depicting 196 tourism landmarks, auto-annotated with SfM tools.
Hyper-parameters are tuned using the IMC validation set. 
As in prior work~\cite{lindenberger2023lightglue,he2023dfsfm}, some scenes of MegaDepth are excluded from training due to their low quality or due to overlap with the IMC test set.
For Co3Dv2, we conduct training and evaluation on 41 categories as in~\cite{wang2023posediffusion, zhang2022relpose, lin2023relposepp}.

We chose a multi-stage training strategy for better stability.
We first train the tracker $\tracker$ on the synthetic Kubric dataset~\cite{greff2021kubric} following the training protocol of \cite{doersch2022tapvid,karaev2023cotracker}.
Then, the tracker is fine-tuned solely on Co3D or MegaDepth, depending on the test dataset. 
Subsequently, we train the camera initializer, with the tracker frozen.
Next, the triangulator is trained with the aforementioned two components held frozen. 
Finally, all components are trained end-to-end. 
In all stages, we randomly sample a training batch of 3 to 30 frames.

\paragraph{Testing.} 
Given input test frames $\setI$, we first select the query frame by identifying the image that is closest to all others based on the cosine similarity between global descriptors extracted by an off-the-shelf ResNet50~\cite{he_deep_2015}. 
Then, we extract SuperPoint and SIFT keypoints from the query frame to serve as the query points for the tracker $\tracker$.
Although our method can track any query point, it performs better when the queries are distinctive.
To improve accuracy, we iterate the whole reconstruction function $f_\theta$ multiple times until reaching sub-pixel BA reprojection error $\mathcal{L}_{\badj}$.
After the first iteration, the query image for each subsequent iteration is the one that is farthest from the query image of the previous iteration (as measured by the ResNet50 cosine similarity).
The re-projections of the point cloud from the current iteration initialize the tracking in the next iteration.


\subsection{Camera pose estimation}

\begin{table}[tbp]
    \centering
    \small
    \resizebox{\linewidth}{!}{
    \begin{tabular}{ccccc} 
    \toprule
      IMC Dataset & Method & AUC@3$\degree$       & AUC@5$\degree$       & AUC@10$\degree$  \\ 
    \midrule
    \multirow{2}{*}{Deep} & DeepSFM & 10.27 & 19.36 & 31.35 \\
    &  PoseDiffusion & 12.31 &  23.17 & 36.82 \\
    \midrule
     \multirow{5}{*}{\shortstack{Incremental\\SfM}}
     &   COLMAP~(SIFT+NN) & 23.58 & 32.66 & 44.79 \\
    &   PixSfM (SIFT + NN) & 25.54 & 34.80 & 46.73 \\
     &  PixSfM (LoFTR) & 44.06 & 56.16 & 69.61 \\
    &   PixSfM (SP + SG) & {45.19}  & 57.22 & 70.47 \\
    & DFSfM~(LoFTR)  & \textbf{46.55}  & \underline{58.74} & \underline{72.19} \\
    \hline
    \multirow{2}{*}{Deep} &  Ours w/o Joint  & 38.23 & 51.60 & 68.35 \\
    & Ours  & \underline{45.23} & \textbf{58.89} & \textbf{73.92} \\
    \bottomrule
    \end{tabular}
    }
    \caption{\textbf{Camera Pose Estimation on IMC.} 
    Our method achieves better accuracy than state-of-the-art Incremental SfM approaches on 2 out of 3 AUC thresholds.
    }
    \label{tab:IMCpose}
\end{table}

Following~\cite{wang2023posediffusion,lindenberger_pixel-perfect_2021,he2023dfsfm,jin2021image}, we report the metric area-under-curve (AUC) to evaluate camera pose accuracy.
In Co3D, similar to~\cite{wang2023posediffusion}, we also report the relative rotation error (RRE) and relative translation error (RTE).
More specifically, for every pair of input frames, we compute the angular translation and rotation error, which are later compared to a threshold yielding accuracies RTE and RRE respectively. 
For a range of thresholds, AUC picks the lower between RRE and RTE, and outputs the area under the accuracy-threshold curve. 
The results on IMC and CO3D are presented in \cref{tab:Co3Dpose} and \cref{tab:IMCpose} respectively. 
For a fair comparison on IMC, we finetune DeepSFM~\cite{wei2020deepsfm} and PoseDiffusion~\cite{wang2023posediffusion} on MegaDepth using their open-source code. The results of Incremental SfM methods are copied from Detector-free SfM~\cite{he2023dfsfm}.

Results indicate that \method outperforms existing methods by a large margin (+9 accuracy points for each metric) on the CO3D dataset.
Here, traditional SfM pipelines suffer because of the wide baselines between test frames.
On IMC, with a good overlap between views, traditional SfM remains superior to recent data-driven deep-learning pipelines \cite{wei2020deepsfm,wang2023posediffusion}.
Our end-to-end trained \method, however, {outperforms all other methods on AUC@10 and AUC@5}, and ranks second on AUC@3, convincingly demonstrating its ability to perform well in both narrow- and wide-baseline regimes.
The accuracy and completeness of our point clouds can be further qualitatively evaluated in \cref{fig:denserec}.

\subsection{3D triangulation}
We evaluate the accuracy and completeness of 3D triangulation on the ETH3D dataset~\cite{schops2017multi} using the same protocol as~\cite{dusmanu2020multi, lindenberger_pixel-perfect_2021, he2023dfsfm}, which triangulates points with fixed camera poses and intrinsics.
Results are shown in \cref{tab:eth3dtriangulation}, where metrics are averaged over all scenes.
Our \method achieves better accuracy and completeness than all baselines (PatchFlow~\cite{dusmanu2020multi}, PixSfM~\cite{lindenberger_pixel-perfect_2021}, and DFSfM~\cite{he2023dfsfm}), regardless of which keypoint detection or matching method they use. 
This is especially obvious from the completeness attained at the 5cm threshold, with our 33.96\% compared to 29.54\% of the best prior work.

\begin{table}[t]
    \centering
    \resizebox{1.0\columnwidth}{!}{
    \setlength\tabcolsep{6pt} %
    \begin{tabular}{ccccccc} 
    \toprule
    \multirow{2}{*}{Method}         & \multicolumn{3}{c}{Accuracy~($\%$)}             & \multicolumn{3}{c}{Completeness~($\%$)} \\ 
    \cmidrule(lr){2-4}
    \cmidrule(lr){5-7}
                            & 1cm       & 2cm       & 5cm       & 1cm       & 2cm       &  5cm \\ 
    \midrule
   PatchFlow (LoFTR)  & 66.73 & 78.73 & 89.93 & 3.48 & 11.34 & 30.96 \\ 
   PixSfM (LoFTR)  & 74.42 & 84.08 & 92.63 & 2.91 & 9.39 & 27.31 \\
   PixSfM (SIFT + NN) & 76.18 & 85.60 & 93.16 & 0.17 & 0.71 & 3.29\\
    PixSfM (SP + SG) & {79.01} & 87.04 & 93.80 & 0.75 & 2.77 & 11.28 \\
    DFSfM (LoFTR) &  \underline{80.38} & \underline{89.01} & \underline{95.83} & \underline{3.73} & \underline{11.07} & \underline{29.54} \\ 
    \midrule
Ours  & \textbf{80.62} & \textbf{89.49} & \textbf{96.52} & \textbf{4.52} & \textbf{13.11} & \textbf{33.96} \\
    \bottomrule
    \end{tabular}
    }
    \vspace{0.15cm}
    \caption{\textbf{3D Triangulation on ETH3D~\cite{schops2017multi}}
    reporting the accuracy and completeness metrics at different thresholds. 
    }
    \label{tab:eth3dtriangulation}
    \end{table}

\subsection{Ablation study}


\paragraph{End-to-end Training.} As reported in~\cref{tab:Co3Dpose} and~\cref{tab:IMCpose}, the end-to-end joint training of the whole framework is important for achieving state-of-the-art performance.
Specifically, comparing \method to an ablation which lacks end-to-end training (Ours w/o joint) we record an improvement from 70.7\% AUC@30 to 74.0\% on the Co3D dataset, and 68.35\% AUC@10 to 73.92\% on the IMC dataset.
This demonstrates the benefits of our fully-differentiable design, and the synergy between its components.

\vspace{2mm}

\paragraph{Tracking or Pairwise Matching.} 
We compare the performance of our predicted tracks to pairwise matching methods on the IMC dataset. 
Specifically, we split our 2D tracks into pairwise matches and feed these matches to PixSfM. 
Also, we construct 2D tracks by pairwise matching (based on the open-source implementation of PixSfM) and feed them to our framework.
It is worth noting that tracks from pairwise matching have a lot of ``holes'' because pairwise matching cannot guarantee proper point tracking.
We fix these holes by setting their locations as the point locations in the query frame, marking them as invisible. 
At the same time, for fair comparison, cameras are still initialized using our tracks, because SP+SG cannot provide track features to our camera initializer.
The results are shown in~\cref{tab:ab_track}.
Although COLMAP (the basis of PixSfM) is designed and carefully engineered for pairwise matching, our tracks achieve a slightly better result than the state-of-the-art matching option SP+SG.
Instead, directly feeding SP+SG tracks to our framework leads to a performance drop.
We attribute this to the fact that using SP+SG tracks cannot benefit from the joint training.
We also provide a qualitative evaluation of our tracking accuracy in~\cref{fig:track_qual} on the IMC dataset.


\begin{table}[t]
    \centering
    \small
    \resizebox{1.0\columnwidth}{!}{
    \begin{tabular}{ccccc} 
    \toprule
    & PixSfM(SP + SG) & PixSfM(Our Tracks) & Ours(SP + LG) & Ours \\
    \midrule
    AUC@10$\degree$ & 70.47 & 70.62 & 68.78 & 73.92   \\
    \bottomrule
    \end{tabular}
    }
    \caption{\textbf{Tracking or Pairwise Matching.} We respectively provide tracks predicted by our tracker or matches estimated by SP+SG to PixSfM and to our \method.
    }\label{tab:ab_track}
    
\end{table}

\paragraph{Camera Initializer and Triangulator.} We also validate the design of our camera initializer $\tformer_\setP$ and triangulator $\tformer_X$ on the IMC dataset.
As reported in \cref{tab:ab_poseinit}, AUC@10 drops clearly if we replace them with alternatives, proving that they provide sufficiently accurate initialization for our global bundle adjustment $\badj$, without the need for incremental camera registration.

\begin{table}[t]
    \centering
    \resizebox{0.8\columnwidth}{!}{
    \scriptsize
    \begin{tabular}{c|cc} 
    \toprule
   &  PoseDiffusion &  Our Camera Initializer   \\
   \midrule
   DLT & 62.18 & 69.42 \\
   Our Triangulator & 66.37 & 73.92 \\
    \bottomrule
    \end{tabular}
    }
    \caption{\textbf{Ablation Study for Camera Initializer and Triangulator.} A clear performance drop is observed when replacing our camera initializer by deep camera prediction method PoseDiffusion, or replacing triangulator by DLT. 
    }\label{tab:ab_poseinit}
\end{table}

\paragraph{Coarse-to-fine Tracking.} 
%
As dicussed above, accurate correspondences are important for structure from motion.
%
We demonstrate the significance of our coarse-to-fine tracking mechanism for the method performance.
By conducting an ablation study where the fine tracker is removed, we observe a significant performance drop on the IMC dataset, with AUC@10 dropping from 73.92\% to 62.30\%.

%% file: sec/5-conclusion.tex
\section{Conclusion}
In this paper, we have presented \method, a fully differentiable SfM approach.
We find that even long-standing pipelines, such as Structure-from-Motion, benefit from a learned adaptation between their components.
This allows \method to be simpler than traditional SfM frameworks while achieving better performance across benchmark datasets.
Moreover, our framework is fully implemented in Python, which will allow for easy modification and improvements in the future.
While \method already achieves good performance, it cannot yet compete with established pipelines in all application domains. 
For example, it currently lacks the capability to process thousands of images as in traditional SfM frameworks. 
%
Nonetheless, we find differentiable SfM a promising direction of research, and our approach lays the foundation for further advances.

%% file: sec/X_suppl.tex
\newpage

\appendix



\section{Implementation Details}

\paragraph{Training} As discussed in the main manuscript, the training process involves multiple stages. We first train the tracker $\tracker$ on the synthetic Kubric dataset, then separately train the tracker $\tracker$, camera initializer $\tformer_{\setP}$, and triangulator $\tformer_X$ on Co3D or MegaDepth, and finally jointly train the whole framework on Co3D or MegaDepth. 
We use the AdamW~\cite{loshchilov2017decoupled} optimizer with a cyclic learning rate scheduler~\cite{smith2019super} where each cycle spans $30$ epochs. 
The learning rate is $0.0001$ for the joint training phase and $0.0005$ for all prior stages. 
We train the model on $32$ NVIDIA A100 ($80$GB) GPUs until convergence. 
The batch size varies for each iteration because we randomly sample $3$ to $30$ frames for each scene (batch) as in~\cite{wang2023posediffusion}. 
The training on the synthetic Kubric dataset takes about one day.
The separate training of the tracker $\tracker$, camera initializer $\tformer_{\setP}$, and triangulator $\tformer_X$ takes two days, two days, and one day respectively. 
The final joint training takes one day. 
For training, we track $256$ query points and run bundle adjustment for $5$ optimization steps. 
We use gradient clipping to ensure stable training, which constrains the gradients' norm to a maximum value of $1$.
Additionally, we normalize the ground-truth cameras in the same way as in \cite{wang2023posediffusion}, and the point cloud correspondingly. 
%

Moreover, we augment the samples using a combination of augmentation transformations. This includes color jittering (brightness, contrast, saturation, and hue) with a $65\%$ probability, Gaussian Blur with a $50\%$ probability, and a $15\%$ chance of converting images to grayscale. 
Please note that different frames from a single scene will receive different augmentations.
Images are resized to $512\times512$ with zero padding. 
%
Ground truth tracks that remain invisible in over $50\%$ of the frames are excluded from the training for the tracker $\tracker$.
For the MegaDepth dataset, similar to ~\cite{sarlin2020superglue, lindenberger2023lightglue}, we construct the training batches by only sampling frames with an overlap score with the query frame exceeding $0.1$.
Here, overlap scores are derived from the pre-processing steps outlined in \cite{dusmanu2019d2}.

\paragraph{Inference Time} On a single NVIDIA A100 $80$GB GPU, given $25$ frames and $4096$ query points, the inference of the tracker, camera initializer, and triangulator takes around $4.3$, $0.9$, and $0.2$ seconds respectively. 
In comparison, the popular pairwise matching variant SuperPoint + SuperGlue usually takes around $20$ seconds. 
In the bundle adjustment process, each optimization step requires approximately $0.7$ seconds. For each run of the whole reconstruction function $f_\theta$ (as discussed in the main manuscript, $f_\theta$ is run multiple times until reaching sub-pixel BA reprojection error), bundle adjustment is executed for $30$ steps, unless early convergence is achieved.

\paragraph{Tracker} We use the 2D convolutional architecture from \cite{karaev2023cotracker, harley2022particle} as the backbone of our tracker. Specifically, for the coarse tracker, this structure consists of an initial convolutional layer with a $7\times7$ kernel and stride of $2$, followed by eight residual blocks with $3\times3$ kernels and instance normalization. 
Finally, the architecture concludes with a pair of convolutional layers, one using a $3\times3$ kernel and the other a $1\times1$ kernel. 
%
This backbone outputs a $128$-dimensional feature map reducing the spatial resolution by a factor of $8$.
We use $5$ levels of correlation pyramids where each level uses a correlation radius of $4$. 
Therefore, the tokens (flattened cost volume) $V \in \mathbb{R}^{N_T \times N_I \times C}$ have a feature dimension of $C = 5 \times (2\times4 + 1)^2 = 405$.
The tokens are subsequently processed by a transformer with eight self-attention layers with a hidden dimension of $512$ and $8$ heads. 
Finally, a multilayer perceptron (MLP) is applied to predict the point location $y$, visibility $v$, and inverse confidence $\sigma$.
The architecture uses GELU activation functions.

The architecture of the fine tracker is similar to the coarse tracker but shallower.
The backbone of the fine tracker consists of one $3\times3$ convolution layer, two residual blocks with $3\times3$ kernels and instance normalization, and one $1\times1$ convolution layer. The correlation pyramid of the fine tracker uses $3$ levels and each level uses a radius of $3$, which leads to tokens with a feature dimension of $3 \times (2\times3 + 1)^2 = 147$. The shallow transformer uses four self-attention layers, with a hidden dimension of $384$ and $4$ heads. 

Following \cite{karaev2023cotracker, harley2022particle}, we train the tracker with $4$ iterative updates and evaluate it with $6$ iterative updates.

\begin{figure}[tbp] 
\includegraphics[width=1.0\linewidth]{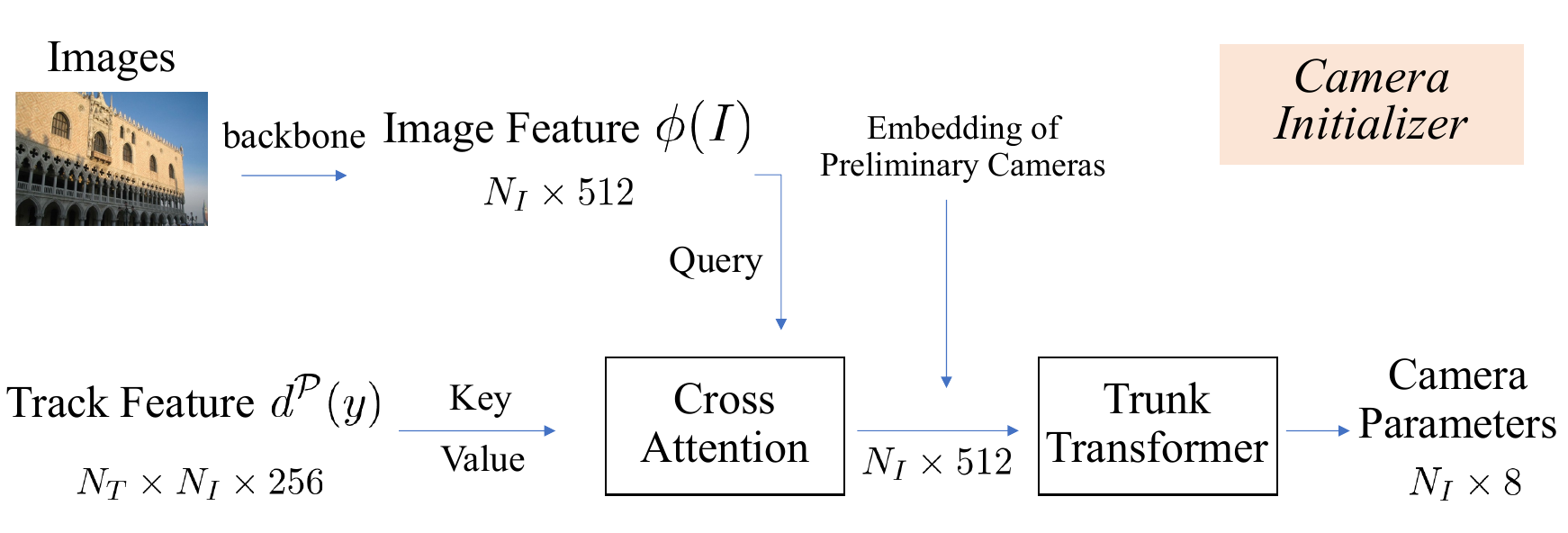}%
\caption{\textbf{Architecture of Camera Initializer.} Generally, we use the image features, track features, and the harmonic embedding of preliminary cameras to predict camera parameters. These parameters, represented in an $N_I \times 8$ matrix, comprise a quaternion ($4$ dimensions), a translation vector ($3$ dimensions), and a focal length ($1$ dimension). 
}\label{fig:camera_arch}
\end{figure}

\paragraph{Camera Initializer} The camera initializer (\cref{fig:camera_arch}) takes frames $I$ and track features $d^{\setP}$ as input, and outputs initial cameras $\hat{\setP}$.
We extract features from the input images in a multi-scale manner as in \cite{wang2023posediffusion}.
However, we use ResNet~\cite{he2016resnet} instead of DINO~\cite{caron2021dino} as the camera initializer backbone, because we empirically found that DINO is harder to train jointly with other components.
Each image is mapped to a $512$-dimensional feature vector $\phi(I_i)$.
Since the track features carry information about the image-to-image correspondence which provides grounding for camera-pose estimation, we fuse the stack of track features $d^{\setP}(\y)$, with shape $N_T \times N_I \times 256$,
into the $N_I \times 512$ image features $\phi(I)$ with $4$ cross-attention layers with $4$ heads.
This results in a $N_I \times 512$ global image descriptor. 

Similar to the tracker, we adopt an iterative update mechanism inside the camera initializer.
For each update, we obtain a set of $8$-dimensionsal preliminary camera representations and map them to $128$ dimensions with a positional harmonic embedding \cite{mildenhall_nerf_2020}.
We then concatenate the global image descriptors and the embedding of the preliminary cameras, use an MLP to project the concatenated features to $512$ dimensions, and feed the latter to a trunk transformer. 
The trunk transformer consists of $8$ self-attention layers (transformer encoder) with $4$ heads, whose hidden dimension is $512$. 
The trunk transformer's output is further processed with another MLP layer, which predicts the camera parameters.
This procedure is repeated four times.
In the first run, the preliminary cameras are derived from each frame's relative camera pose to the query frame, which is computed from tracks using the 8-point algorithm.   
Following the approach of COLMAP~\cite{schoenberger2016sfm}, the focal lengths are initialized based on the longer side of the image size.
In subsequent runs, the preliminary cameras (intrinsic and extrinsic) are the result of the previous prediction. 
In this process, the trunk transformer is run four times while the feature backbone is only run once.

It is noteworthy that the traditional 8-point algorithm is commonly used in conjunction with RANSAC to filter out noisy matches. 
In our approach, we employ a batched 8-point algorithm to approximate a similar effect to RANSAC while avoiding a time-consuming \textit{for} loop. 
For each scene, we randomly select $20$ sets, each comprising $50$ point pairs. 
We then apply the 8-point algorithm to these sets in parallel, yielding $20$ relative camera poses. 
Similar to RANSAC, we calculate the inlier count for each camera pose candidate using all available point pairs. 
%
%
A point pair is considered as an inlier if its Sampson epipolar error is less than $0.6$ divided by the image width in pixels.
Ultimately, the camera pose candidate with the highest number of inliers is selected. 
Our implementation of the 8-point algorithm is based on \cite{wei2023generalized}.

\begin{figure}[tbp] 
\includegraphics[width=1.0\linewidth]{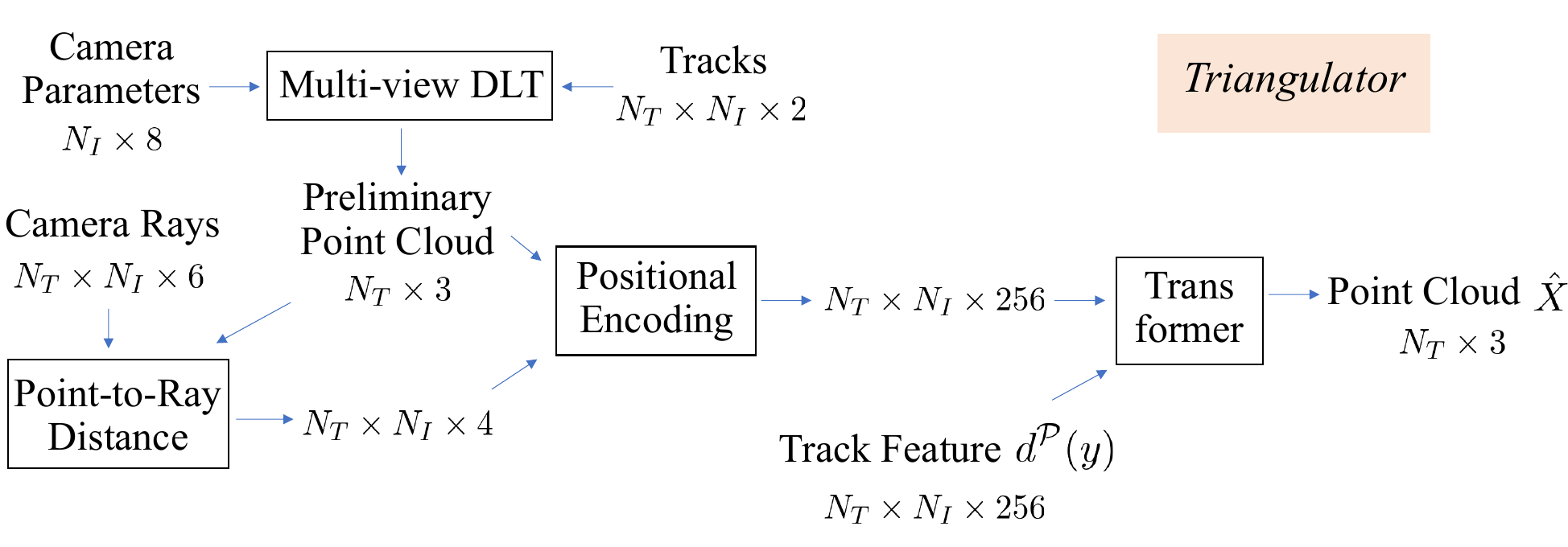}%
\caption{\textbf{Architecture of Triangulator.} We first estimate a preliminary point cloud using the camera parameters and tracks. 
%
Subsequently, we calculate the distance from this preliminary point cloud to all camera rays, as well as identify the nearest points on these rays. 
This information (along with the preliminary point cloud) is concatenated to the track features and fed into a transformer to predict the point cloud $\hat{X}$.
}\label{fig:triangulator_arch}
\end{figure}

\paragraph{Triangulator} 
Given camera parameters and tracks, the triangulator (\cref{fig:triangulator_arch}) $\tformer_X$ initially estimates a preliminary point cloud $\bar{X}$ (of size $N_T \times 3$) using a closed-form multi-view Direct Linear Transform (DLT) for 3D triangulation.
Furthermore, for each frame and the corresponding 2D point, a camera ray is computed.
The distance from this camera ray to the associated 3D point in $\bar{X}$, along with the nearest point on the camera ray, are calculated.
This results in the preliminary point cloud $\bar{X}$ with shape $N_T \times 3$, the ray distance with shape $N_T \times N_I \times 1$ and nearest points to camera rays of shape $N_T \times N_I \times 3$.
These vectors are then concatenated (resulting in a tensor of shape $N_T \times N_I \times 7$) and embedded into a 256-dimensional space ($N_T \times N_I \times 256$) through positional encoding. 
The embedded vectors are further concatenated with the track feature $d^{\setP}(\y)$, leading to a shape of $N_T \times N_I \times 512$. 
Averaging over the $N_I$ dimension yields a descriptor for the point cloud with dimensions $N_T \times 512$. This descriptor is input into a transformer comprising $4$ self-attention layers, each with $4$ heads and a hidden dimension of $384$. 
The output of the transformer is processed by a two-layer MLP (the hidden dimension is $256$) to estimate $\hat{X}$.

\paragraph{Outlier Filtering} It is important to filter out noisy correspondences in SfM, especially for BA optimization. For our framework, first, we drop 2D points with a visibility score $v<0.6$ or variance $\sigma > 1$ (horizontally or vertically). 
Then, we use the preliminary cameras estimated by the 8-point algorithm and the initial cameras $\hat{\setP}$ to remove correspondences with a Sampson epipolar error of more than $0.8$ divided by the image width.
Following Bundler \cite{snavely2006photo} and COLMAP\cite{schoenberger2016sfm}, we also require that at least one pair within each track has a triangulation angle of more than $3$ degrees. 
Otherwise, the track (and the associated 3D point) is discarded. 
Moreover, for bundle adjustment, the 2D points with a reprojection error of more than $3$ pixels are removed. 
Tracks with less than 3 points are discarded as well.
It is worth mentioning that Homography verification~\cite{schoenberger2016sfm} does not seem to be important for our framework, although it is common in incremental SfM. 

\section{Discussions and Ablation}


\paragraph{Global SfM} As discussed in the Related Work section of the main manuscript, there are two popular approaches for SfM: incremental and global. Global SfM approaches \cite{arie2012global, crandall2012sfm, cui2015linear, moulon2013global, sweeney2015optimizing, rother2003linear, ozyesil2015robust, jiang2013global, cui2015global, cui2017hsfm} usually predict the parameters for all the cameras at the same time and only perform bundle adjustment once. 
These methods often use rotation averaging and translation averaging to align pairwise relative camera poses into a consistent coordinate system. 
Our proposed method bears similarities to global SfM. However, it diverges in several key aspects: (1) unlike global SfM, which relies on pairwise matching (akin to incremental SfM), our method directly predicts tracks; (2) instead of rotation averaging and translation averaging in global SfM, we use a learnable network to predict camera parameters; (3) we iteratively apply the reconstruction function multiple times during testing, with bundle adjustment at each iteration. 
Besides these differences, our method is complementary to global SfM. 

\begin{table}[tbp]
    \centering
    \small
    \begin{tabular}{c|ccc} 
    \toprule
   &  w/o Filtering &  w/o BA & Ours   \\
   \midrule
    AUC@10$\degree$ & 2.31 & 18.34 &  73.92 \\
    RRE@5$\degree$ & 8.17 & 70.25 & 95.61 \\
    RTE@5$\degree$ & 5.42 & 39.42 & 81.03 \\
    \bottomrule
    \end{tabular}
    \caption{\textbf{Ablation Study for Bundle Adjustment.} We try the setting without using bundle adjustment, or using bundle adjustment but not filtering the correspondences. 
    }\label{tab:ab_ba}
\vspace{-4mm}
\end{table}

\paragraph{Bundle Adjustment}  Bundle adjustment is a key component for accurate SfM.
As shown in \cref{tab:ab_ba}, without bundle adjustment, we observe a clear performance drop, with AUC@10 from $73.92$ to $18.34$.
BA is also known to be strongly susceptible to noisy inputs.
Indeed, using bundle adjustment without track filtering (described in previous paragraphs) destroys the estimate and, as such, reduces the AUC@10 nearly to zero. 
%
Notably, even without bundle adjustment, our framework's estimation remains relatively robust; for instance, the rotation errors for over $70\%$ of image pairs remain within $5$ degrees (RRE@5$\degree$ $> 70\%$).
However, executing bundle adjustment without track filtering results in incorrect optimization of camera parameters, whose RRE@5$\degree$ is also just around $8\%$. 
At the same time, please note that all the methods in Table 2 of the main manuscript use bundle adjustment or its approximation. 
For example, PoseDiffusion~\cite{wang2023posediffusion} uses geometry-guided sampling and DeepSfM~\cite{wei2020deepsfm} adopts a special form of bundle adjustment.
Without geometry-guided sampling, the AUC@10 of PoseDiffusion is around $11\%$.



\begin{figure}[htbp] 
\includegraphics[width=1.0\linewidth]{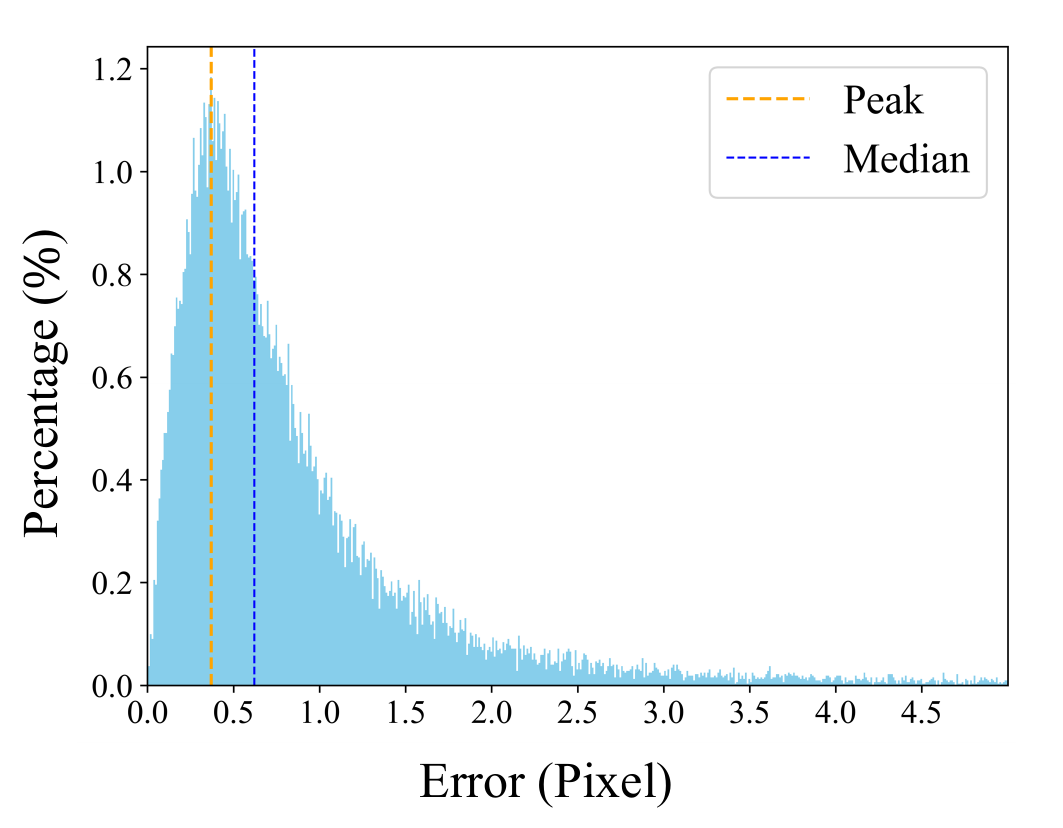}%
\caption{\textbf{Histogram of Tracking Errors} of the scene \textit{British Museum 10 bag 000} on the IMC dataset. The horizontal axis denotes {error in pixels}, while the vertical axis shows the {percentage (\%)} for each bin.
}\label{fig:track_hist}
\end{figure}

\paragraph{Track Error Distribution} We present a histogram in \cref{fig:track_hist} to depict the distribution of tracking errors for the scene \textit{British Museum 10 bag 000} within the IMC dataset, consisting of 10 images. As indicated, the distribution's peak, represented by the orange dashed line, approximately aligns with $0.4$ pixels, while the median, depicted by the blue dash-dotted line, is around $0.6$ pixels.   Notably, most of the tracking predictions maintain an error margin of less than $3$ pixels, highlighting the accuracy of our method.
Some predictions even approach a near-zero error margin.
Invisible points (\eg, occluded or outside the view) are not included in this histogram.

\begin{table}[tbp]
    \centering
    \small
    \begin{tabular}{c|ccc} 
    \toprule
   &  AUC@10$\degree$ &  RRE@5$\degree$ & RTE@5$\degree$   \\
   \midrule
    PiPs~\cite{harley2022particle} & 43.27 & 82.15 & 51.39 \\
    Our Trakcer & 73.92 & 95.61 &  81.03  \\
    \bottomrule
    \end{tabular}
    \caption{\textbf{Ablation Study for Tracking.} We try the video tracking method PiPs~\cite{harley2022particle} in our framework, which shows a clear performance drop .
    }\label{tab:ab_tracking}
\vspace{-4mm}
\end{table}

\paragraph{Video Tracking} To verify the effect of our proposed tracker, we also try to use the video tracking method PiPs inside our framework. 
The results, presented in Table \ref{tab:ab_tracking}, reveal a noticeable decline in performance when using PiPs as opposed to our tracker.
This contrast underlines the effectiveness of our proposed tracking solution.